\documentclass[runningheads]{llncs}
\usepackage{graphicx}
\usepackage{amsmath,epsfig}
\usepackage{cite}
\usepackage{siunitx}
\usepackage{amsmath,amssymb,amsfonts}
\usepackage{algorithmic}
\usepackage{textcomp}
\usepackage{tabularx}
\usepackage{subcaption}
\usepackage{hyperref}

\usepackage{booktabs, multirow} 
\usepackage{soul}
\usepackage[table]{xcolor} 
\usepackage{changepage,threeparttable} 
\begin{document}
\title{Analysis and application of multispectral data for water segmentation using machine learning}
\titlerunning{water segmentation using machine learning}
%
\author{Shubham Gupta\inst{1}\thanks{Work done partially while interning at CDSAML, PES University and RRSC-S, ISRO.} \and
Uma D.\inst{1} \and
Hebbar, R. \inst{2}}
\authorrunning{S. Gupta et al.}
%
\institute{PES University, Bangaluru KA 560085, India \and
Regional Remote Sensing Centre - South, ISRO, Bengaluru KA 560037, India
\email{shubhamgupto@gmail.com}}

\maketitle              
\begin{abstract}
Monitoring water is a complex task due to its dynamic nature, added pollutants, and land build-up. The availability of high-resolu-tion data by Sentinel-2 multispectral products makes implementing remote sensing applications feasible. However, overutilizing or underutilizing multispectral bands of the product can lead to inferior performance. In this work, we compare the performances of ten out of the thirteen bands available in a Sentinel-2 product for water segmentation using eight machine learning algorithms. We find that the shortwave infrared bands (B11 and B12) are the most superior for segmenting water bodies. {\textcolor{black}{B11 achieves an overall accuracy of $71\%$ while B12 achieves $69\%$ across all algorithms on the test site.}} We also find that the Support Vector Machine (SVM) algorithm is the most favourable for single-band water segmentation. {\textcolor{black}{The SVM achieves an overall accuracy of $69\%$ across the tested bands over the given test site.}} Finally, to demonstrate the effectiveness of choosing the right amount of data, we use only B11 reflectance data to train an artificial neural network, BandNet. Even with a basic architecture, BandNet is proportionate to known architectures for semantic and water segmentation, achieving a $92.47$ mIOU on the test site. BandNet requires only a fraction of the time and resources to train and run inference, making it suitable to be deployed on web applications to run and monitor water bodies in localized regions. {\textcolor{black}{Our codebase is available at \href{https://github.com/IamShubhamGupto/BandNet}{https://github.com/IamShubhamGupto/BandNet}}}.

\keywords{Water  \and Sentinel-2 \and machine-learning \and artificial neural network \and BandNet}
\end{abstract}
\section{Introduction}
Water is one of the most essential resources on our planet. With the rise in population in cities, water bodies have shrunk or disappeared entirely. In the coastal regions, due to global warming, the increase in water levels has begun to submerge land in low-lying areas\cite{r0}. To take action and prevent irreparable damages, spatio-temporal awareness of water is a necessity. 

The recent advancements in Convolutional Neural Network (CNN) models show promising results in object classification{\textcolor{black}{\cite{dino, swinv2} and segmentation\cite{deeplabv3+,encoder-decoder, swin} from images and reflectance values\cite{water_survey, watnet, deepwatermapv2}}}. The {\textcolor{black}{reflectance}} models rely on the combination of either or all of the visible spectrum, near-infrared, and shortwave-infrared spectrum bands. To our knowledge, there {\textcolor{black}{is}} no ranking of the multispectral bands for their usability for water segmentation. In this work, we study the {\textcolor{black}{relationship}} between water segmentation performance and multispectral bands using several machine learning algorithms\cite{lr, gnb, knn, dt, rf, xgb,  sgd, svm}. The goal is to find the multi-spectral band and machine learning algorithm that is best suited to segmenting water bodies. 

By utilizing the {\textcolor{black}{best multispectral band}} to delineate water bodies, we show that a single band is adequate to perform similar to large segmentation models\cite{deeplabv3+} trained on multispectral false-infrared images {\textcolor{black}{ that are composed of B8, B4, and B3. We achieve this result on a simple Artificial Neural Network, BandNet.}} Unlike existing literature, BandNet does not treat multispectral data as images but rather sees each pixel as a unique data-point. Furthermore, it requires a fraction of the training time and resources to produce respectable results. To summarize our contributions to this paper:
\begin{enumerate}
\item We compare the water segmentation performance of the multispectral bands present in the Sentinel-2 product using multiple machine learning algorithms.
\item We compare the performances of various machine learning algorithms for single-band water segmentation.
\item {\textcolor{black}{We demonstrate the importance of data selection by training a simple ANN, BandNet, that performs similar to other segmentation models - trained on images and relfectance data.}} 
\end{enumerate}

\section{Satellite Data and Study Site}

This study uses the multi-spectral products from the European Space Agency satellite constellation Sentinel-2. All the products are made available in L2A mode, giving us direct access to Bottom of Atmosphere (BOA) reflectance\cite{bao}. We study bands B2, B3, B4, B5, B6, B7, B8, B8A, B11, and B12 which are re-sampled to $10m$ spatial resolution {\textcolor{black}{using SNAP, SeNtinel Applications Platform}}.

Since one of the requirements of training Deeplabv3+\cite{deeplabv3+} is an extensively annotated dataset, we required images of water bodies from more than one Sentinel-2 product. For this reason, we acquired a total of eight Sentinel-2 products. As for the image from the products themselves, we utilize false infrared images generated from B8, B4, and B3 of each product. The entire image is 10980x10980 pixels. Since we cannot process an image of this scale at once, we break down the false-infrared images into 549x549 pixel tiles.

\begin{table}[hp]
\caption{Metadata of the Sentinel-2 data used. \textsuperscript{*}Rotation is applied to images as part of data augmentation.}\label{tab:data}
\begin{tabularx}{\textwidth}{|p{25mm}|p{25mm}|p{42mm}|p{25mm}|}
\hline
\textbf{Tile}	&\textbf{Date}	&\textbf{Type} &\textbf{Images}\\
\hline
T43PGQ				&2019-01-04		&Lakes      &186\textsuperscript{*}\\
T44RQN				&2019-01-21		&Rivers     &64\\ 
T43PFS				&2019-02-16		&Lakes, Rivers      &47\\ 
T43QGU				&2019-03-18 	&Lakes, Rivers      &172\textsuperscript{*}\\ 
T43PFQ				&2019-03-28		&Lakes      &310\textsuperscript{*}\\
T44QLE				&2019-03-30		&Rivers     &175\\
T43RFJ				&2019-04-20		&Rivers     &242\textsuperscript{*}\\ 
T43QHV				&2019-04-24 	&Lakes      &250\textsuperscript{*}\\
\hline
\end{tabularx}
\end{table}
We use the scene classification map available within each Sentinel-2 product to annotate water bodies. We discard tiles and the corresponding annotation if they do not contain any water bodies. The resulting data acquired from these products is summarized in Table \ref{tab:data}. We obtain 1446 images with annotations after applying a simple rotation augmentation to five of the eight products.

\begin{figure}[htbp]
\captionsetup[subfigure]{justification=centering}
  \centering
  \begin{subfigure}[b]{0.39\linewidth}
    \includegraphics[scale = .7]{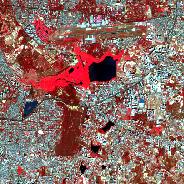}
    \caption{}
  \end{subfigure}
  \begin{subfigure}[b]{0.39\linewidth}
    \includegraphics[scale = .7]{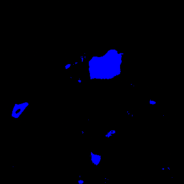}
    \caption{}
  \end{subfigure}
  \begin{subfigure}[b]{0.39\linewidth}
    \includegraphics[scale = .7]{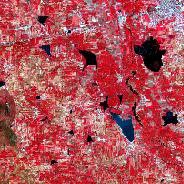}
    \caption{}
  \end{subfigure}
  \begin{subfigure}[b]{0.39\linewidth}
    \includegraphics[scale = .61]{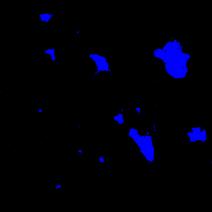}
    \caption{}
  \end{subfigure}
  \caption{False infrared image generated from the Sentinel-2 product T43PGQ subset dated 2019-01-04. \textbf{(a)}Preview of subset1.  \textbf{(b), (d)} Corresponding water body annotation generated from scene classification map. \textbf{(c)}Preview of subset2.} 
 \label{fig:previewimg}
\end{figure}
On the other hand, a single Sentinel-2 product can provide $120$ million data points as the values are based on each pixel. Due to hardware limitations, we can only take a subset of the product at a time. Figure\ref{fig:previewimg} shows the false-infrared images of the two subsets we will be using to evaluate our findings. Image (a) in Figure\ref{fig:previewimg}  represents a subset with North latitude 12.963, West longitude 77.630, South latitude 12.889, East longitude 77.703. This subset contains Bellandur lake and has a good representation of land built-up, water bodies, and vegetation. Image (c) in  Figure\ref{fig:previewimg}  represents a subset with North latitude 12.655, West longitude 77.131, South latitude 12.581, East longitude 77.205. This subset contains vegetated land and lakes at close proximity. We will be referring to them as subset1 and subset2 respectively.

\section{Methodology Used}
\label{methodology}
\subsection{Data processing}


From subset1, we consider an equal number of water and non-water data points. For reproducibility, we set a seed value\cite{seed} so that the same data points will be selected when the experiment is re-run. The dataset is split into test, train, and validation sets, each being mutually exclusive. We evaluate and report the performance of the machine learning algorithms on the validation set. The validation set has never been seen before by the algorithms. Data points from subset2 are entirely used to evaluate the algorithms and models. 
\subsection{Band reflectance analysis}
We employ eight statistical machine learning algorithms on the reflectance data of individual bands to gauge their performance for water segmentation, namely: Logistic Regression (LR)\cite{lr}, Gaussian Naive Bayes (GNB)\cite{gnb}, K-neighbours (KN)\cite{knn}, Decision Tree (DT)\cite{dt}, Random Forest (RF)\cite{rf}, XGBoost (XGB)\cite{xgb}, Stochastic Gradient Descent (SGD)\cite{sgd}, and Support Vector Machine (SVM)\cite{svm}. {\textcolor{black}{All eight algorithms are first made to fit the training set, followed by testing their performance on the validation set.}}

\subsection{BandNet}
Based on the performance of individual bands, further discussed in Section\ref{results}, we designed a simple ANN architecture, BandNet, to create segmentation maps of water bodies using raw reflectance data as input.
\begin{figure}[hb]
\centerline{\includegraphics[scale=0.32]{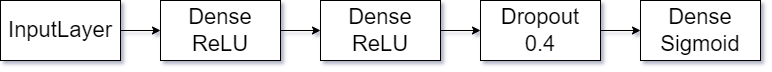}}
\caption{Proposed architecture of BandNet for segmentation of Water Bodies.}
\label{fig:bandnet-arch}
\end{figure}

\noindent BandNet consists of two Dense layers with ReLU activation functions. Before the classification head, a Dropout layer of 0.4 is in place to deal with the class imbalance problem and improve generalization. The classification head is a Dense layer with Sigmoid activation. The architecture of BandNet is shown in Figure\ref{fig:bandnet-arch}.
\subsection{Multi-spectral Image Analysis}
We generate false colour infrared images from B8, B4, and B3 to train Deeplabv3+\cite{deeplabv3+}, a deep neural network architecture that uses Atrous Spatial Pyramid Pooling (ASPP)\cite{atrous1} and a encoder–decoder\cite{encoder-decoder} structure. The ASPP is used to encode the contextual information with the help of varying atrous rates $r$.
\begin{equation}
y[i] = \sum_{k}x[i +r.k]w[k]
\end{equation}
The decoder then uses the contextual information to generate fine object boundaries. Finally, depthwise separable convolution\cite{xception} is applied to both ASPP and the decoder module for reducing computational complexity while preserving the performance of the model.

\section{Implementation details}
Here we define the details used to train the various models and algorithms. We use an Nvidia GTX 1050 graphics card, 16GB of memory, and an Intel i5 8300H processor for training and inference. The SVM classifier uses a radial basis function (rbf) kernel. The RF classifier uses 100 estimators. The KN classifier uses seven neighbors to vote. The SGD classifier uses a Huber loss function with an L1 penalty for 25 iterations. The remaining machine learning algorithms ran with no extra parameters.

For each multispectral band and model, we generate a confusion matrix, which then helps us calculate the mean intersection over union (mIoU). mIoU for N classes can be defined in terms of true positives (TP) as:

\begin{equation}
IoU\textsubscript{i} = \frac{TP\textsubscript{i}}{Sum Of Rows\textsubscript{i} + Sum Of Columns\textsubscript{i} - TP\textsubscript{i}}    
\end{equation}
\begin{equation}
mIoU = \frac{\sum_{n=1}^{N}	IoU\textsubscript{i}}{N}
\end{equation} 
We train BandNet with a BinaryCrossEntropy loss function and BinaryAccuracy as the metric. The optimizer Adam has a learning rate of $1e-4$. We use EarlyStopping to avoid overfitting with a patience value of 5.

We run the DeepwatermapV2\cite{deepwatermapv2} and WatNet\cite{watnet} directly using pretrained weights for inference on B2, B3, B4, B8A, B11, and B12 based on the guides provided by the authors. 

The Deeplabv3+ uses a learning rate of $0.001$, weight decay of $0.00004$, atrous rates at (6,12,18), output stride of $16$, momentum of $0.9$, decoder output stride at $4$, batch size of $1$ for 30K iterations. We use Xception65\cite{xception} and MobileNetv2\cite{mobilenetv2} as the network backbones pretrained on the PASCALVOC-2012 dataset\cite{pascal-voc-2012}.

\section{Results and Discussion}
\label{results}
\begin{table}[htbp]\centering
\caption{Comparison of single band water segmentation performance (mIoU) using various machine learning models on validation set of subset1. The mIoU is colour coded from low to high as {\color{purple}brick-red} to  {\color{blue}blue}. The Percent row and column describes the accuracy of a band across various algorithms and the accuracy of an algorithm across various bands respectively. The colour coding from low to high follows as {\color{red} red} to {\color{teal} teal}.}\label{tab:compare-ml}
\scriptsize
\begin{tabularx}{0.99\textwidth}{|p{10.5mm}|p{11mm}p{11mm}p{11mm}p{11mm}p{11mm}p{11mm}p{11mm}p{11mm}c|p{11mm}|}\toprule
Band &LR &GNB &RF &KN &DT &SGD &XGB &SVM & &Percent \\\midrule
B02 &\cellcolor[HTML]{a4180b}36.93 &\cellcolor[HTML]{9e0b05}34.23 &\cellcolor[HTML]{b13218}42.07 &\cellcolor[HTML]{b23619}42.74 &\cellcolor[HTML]{b23419}42.45 &\cellcolor[HTML]{a01007}35.26 &\cellcolor[HTML]{b23519}42.62 &\cellcolor[HTML]{ae2c15}40.76 & &\cellcolor[HTML]{e67c73}0.35 \\
B03 &\cellcolor[HTML]{a4180b}36.79 &\cellcolor[HTML]{990000}32.04 &\cellcolor[HTML]{b03017}41.58 &\cellcolor[HTML]{b03117}41.69 &\cellcolor[HTML]{b03217}41.89 &\cellcolor[HTML]{ae2d15}40.99 &\cellcolor[HTML]{af2f16}41.45 &\cellcolor[HTML]{b13218}42.06 & &\cellcolor[HTML]{e67c73}0.35 \\
B04 &\cellcolor[HTML]{b63e1d}44.37 &\cellcolor[HTML]{b23519}42.66 &\cellcolor[HTML]{c7622e}51.34 &\cellcolor[HTML]{c8632f}51.58 &\cellcolor[HTML]{c8642f}51.82 &\cellcolor[HTML]{c05227}48.33 &\cellcolor[HTML]{c7622e}51.43 &\cellcolor[HTML]{df9346}61.06 & &\cellcolor[HTML]{ec9d97}0.43 \\
B05 &\cellcolor[HTML]{c25729}49.17 &\cellcolor[HTML]{c05227}48.23 &\cellcolor[HTML]{d07437}55.02 &\cellcolor[HTML]{d17638}55.3 &\cellcolor[HTML]{d27939}55.86 &\cellcolor[HTML]{d8843f}58.11 &\cellcolor[HTML]{d47c3b}56.51 &\cellcolor[HTML]{d47c3b}56.48 & &\cellcolor[HTML]{f0b0ab}0.47 \\
B06 &\cellcolor[HTML]{facc61}72.26 &\cellcolor[HTML]{c7b870}77.27 &\cellcolor[HTML]{dec46c}76.02 &\cellcolor[HTML]{fbce62}72.68 &\cellcolor[HTML]{d1bd6e}76.75 &\cellcolor[HTML]{a0a377}79.44 &\cellcolor[HTML]{a1a477}79.41 &\cellcolor[HTML]{aaa876}78.92 & &\cellcolor[HTML]{eff9f4}0.67 \\
B07 &\cellcolor[HTML]{fdd364}73.66 &\cellcolor[HTML]{9fa278}79.53 &\cellcolor[HTML]{d5c06e}76.52 &\cellcolor[HTML]{fed465}73.88 &\cellcolor[HTML]{c9b970}77.16 &\cellcolor[HTML]{b1ac74}78.5 &\cellcolor[HTML]{c1b571}77.62 &\cellcolor[HTML]{b7af73}78.18 & &\cellcolor[HTML]{e3f4eb}0.67 \\
B08 &\cellcolor[HTML]{f5c15c}70.15 &\cellcolor[HTML]{efcd69}75.08 &\cellcolor[HTML]{fbd466}74.4 &\cellcolor[HTML]{bab173}78.02 &\cellcolor[HTML]{e3c76b}75.72 &\cellcolor[HTML]{fcd163}73.16 &\cellcolor[HTML]{a7a776}79.05 &\cellcolor[HTML]{9aa079}79.78 & &\cellcolor[HTML]{fefcfc}0.66 \\
B8A &\cellcolor[HTML]{e1c66b}75.81 &\cellcolor[HTML]{648383}82.77 &\cellcolor[HTML]{c9b970}77.16 &\cellcolor[HTML]{fbcf62}72.82 &\cellcolor[HTML]{bfb472}77.72 &\cellcolor[HTML]{b7af73}78.19 &\cellcolor[HTML]{bab173}78.03 &\cellcolor[HTML]{9ca178}79.69 & &\cellcolor[HTML]{ccebdb}0.68 \\
B11 &\cellcolor[HTML]{89977c}80.75 &\cellcolor[HTML]{0b5394}87.71 &\cellcolor[HTML]{778d7f}81.71 &\cellcolor[HTML]{afab75}78.61 &\cellcolor[HTML]{6d8881}82.3 &\cellcolor[HTML]{bab173}78 &\cellcolor[HTML]{6f8981}82.2 &\cellcolor[HTML]{437189}84.65 & &\cellcolor[HTML]{57bb8a}0.71 \\
B12 &\cellcolor[HTML]{f8c75f}71.33 &\cellcolor[HTML]{26628e}86.21 &\cellcolor[HTML]{bab173}78.01 &\cellcolor[HTML]{b6af73}78.21 &\cellcolor[HTML]{959d7a}80.06 &\cellcolor[HTML]{c6b770}77.36 &\cellcolor[HTML]{427189}84.68 &\cellcolor[HTML]{427189}84.69 & &\cellcolor[HTML]{96d5b6}0.69 \\
\hline
& & & & & & & & & & \\
Percent &\cellcolor[HTML]{e67c73}0.61 &\cellcolor[HTML]{fbeeed}0.65 &\cellcolor[HTML]{f0f9f5}0.65 &\cellcolor[HTML]{fbeeed}0.65 &\cellcolor[HTML]{cbeadb}0.66 &\cellcolor[HTML]{fcf4f3}0.65 &\cellcolor[HTML]{96d5b6}0.67 &\cellcolor[HTML]{57bb8a}0.69 & & \\
\bottomrule
\end{tabularx}
\end{table}

\noindent In table\ref{tab:compare-ml}, we present our results of generating water segmentation from each of the ten bands paired with each of the eight machine learning algorithms. We calculate the  percent column for a band as a fraction of the sum of the achieved mIoU across all algorithms, to the maximum attainable mIoU. The percent row for each algorithm is calculated as the sum of the achieved mIoU across all bands divided by the maximum attainable mIoU.

In that order, B11, B12, and B8A are the best-performing bands across all tested algorithms. B11 outperforms B2 by an absolute value of $0.36$ which makes B11 $102\%$ better than B2 for water segmentation. We also notice a trend of SWIR bands (B11 and B12), followed by NIR bands (B8, B8A) and finally visible spectrum bands (B2, B3, B4) in terms of water segmentation performance.
However, the algorithms {\textcolor{black}{used to test these bands}} do not present such large variations in performance. The best performing algorithm, Support Vector Machines\cite{svm}, outperforms Linear Regression\cite{lr} by an absolute value of $0.08$ or $13.11\%$. 

\begin{table}[ht]
\centering
\caption{Comparison of BandNet to other models on subset1 in terms of performance, training time, parameters. mIOU: mean Intersection over Union. Time is measured in Hours. $\downarrow$: The lower the better. $\uparrow$: The higher the better. }
\label{tab:ann-results}
\begin{tabularx}{\textwidth}{|p{32mm}|p{36mm}|p{23mm}|p{12mm}|p{13mm}|}
\toprule
\textbf{Model}  &\textbf{Data}   &\textbf{Parameters $\downarrow$ }     &\textbf{Time $\downarrow$}   &\textbf{mIOU $\uparrow$}\\
\midrule
DeepWaterMapv2\cite{deepwatermapv2} & B2-B3-B4-B8A-B11-B12 & 37.2M &- &89.01\\
WatNet\cite{watnet}     &B2-B3-B4-B8A-B11-B12 &3.4M &- &89.74\\
BandNet (Ours)         &B8-B4-B3    &1217              &0.41       &89.92\\
MobileNetv2\cite{deeplabv3+,mobilenetv2}  &Image         &2.1M           &12       &90.27\\   
Xception65\cite{deeplabv3+,xception}   &Image         &41M            &16       &92.44\\
\textbf{BandNet (Ours)}         &\textbf{B11}         &\textbf{1153}              &\textbf{0.45}       &\textbf{92.47}\\
BandNet (Ours)         &B12-B11     &1185              &0.27       &92.96\\
BandNet (Ours)         &B12-B11-B8A &1217              &0.65       &92.99\\
\bottomrule
\end{tabularx}
\end{table}
\begin{table}[htbp]
\centering
\caption{Comparison of BandNet to reflectance models on subset1 in terms of performance, training time, parameters. mIOU: mean Intersection over Union. Time is measured in Hours. $\uparrow$: The higher the better.}
\label{tab:subset2}
\begin{tabular}{|p{35mm}|p{40mm}|p{15mm}|}
\toprule
\textbf{Model}  &\textbf{Data}   &\textbf{mIOU $\uparrow$}\\
\midrule
WatNet\cite{watnet}     &B2-B3-B4-B8A-B11-B12 &86.92\\
BandNet (Ours)         &B12-B11-B8A &87.00\\
BandNet (Ours)         &B12-B11     &87.33\\
BandNet (Ours)       &B11    &87.42\\
DeepWaterMapv2\cite{deepwatermapv2} & B2-B3-B4-B8A-B11-B12 &89.33\\
\bottomrule
\end{tabular}
\end{table}
In table \ref{tab:ann-results}, we compare the performance of BandNet on different band combinations to existing solutions for water segmentation.{\textcolor{black}{This comparison is carried out on subset1.}} We directly use Deepwatermapv2 and WatNet using their existing weights provided by the authors. Both models perform respectably considering they have never seen the data before in training or testing phases.

To make a fair comparison of images to reflectance data, we compared Deeplab V3+ to BandNet trained on the reflectance of B8-B4-B3. We notice BandNet performs similarly to Deeplabv3+ with MobileNetv2 as the backbone. However, BandNet trains at a fraction of the parameters and time required by Deeplabv3+ for the same performance. This performance gap is bridged when we train BandNet on B11, the best performing multispectral band. We observe that BandNet with B11 outperforms Deeplabv3+ by an absolute value of $2.2$ and $0.03$ mIOU on MobileNetv2 and Xception65 backbones, respectively. BandNet further surpasses Deeplab3+ if we pass along additional high-performing bands with B11.

In table\ref{tab:subset2}, we compare the inference of models on subset2. BandNet sees a slight drop in performance but is still comparable to Deepwatermapv2 and WatNet. {\textcolor{black}{These results underscore the idea behind this work that choosing the right set of data is important when building predictive models.}} We believe this drop in performance is due to the simplistic nature of BandNet and that it was originally designed for one multispectral band only. However, this makes it convenient to re-train and deploy BandNet over a very localized region using minimal resources and time. The small size of BandNet allows it to be hosted on a web application and run inferences at a very low cost. This is extremely helpful if someone were to host their own water body monitoring solution. 
\begin{figure}[htbp]
\captionsetup[subfigure]{justification=centering}
  \centering
  \begin{subfigure}[b]{0.328\linewidth}
    \includegraphics[scale = 0.49]{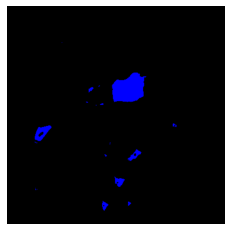}
    \caption{}
  \end{subfigure}
  \begin{subfigure}[b]{0.328\linewidth}
    \includegraphics[scale = 0.49]{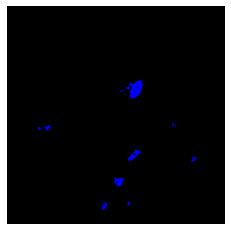}
    \caption{}
  \end{subfigure}
  \begin{subfigure}[b]{0.328\linewidth}
    \includegraphics[scale = 0.49]{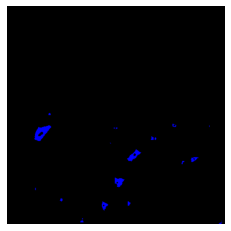}
    \caption{}
  \end{subfigure}
  \caption{Application of BandNet: Monitoring of Bellandur lake using BandNet in subset1. The maps are generated on the T43PGQ subset dated: \textbf{(a)} 2019-01-04, \textbf{(b)} 2020-03-29 and \textbf{(c)} 2021-03-04.}
\end{figure}
\section{Conclusion}
In this paper, we have compared the performances of individual bands for water segmentation, using machine learning algorithms. We create a hierarchy of multispectral bands based on their performance, placing SWIR bands (B11, B12) on the top, followed by NIR bands (B8, B8A) and finally visible spectrum bands (B2, B3, B4). We also observed that the Support Vector Machine followed by XG Boost algorithm are favourable is single band water segmentation compared to other algorithms.

Using the best performing band, B11, we developed a simple ANN that is able to compete with specialized segmentation architectures in performance while requiring only a fraction of the time and resources to train. This lightweight nature of BandNet makes it suitable for deploying it on web applications to monitor water bodies in localized regions.

The objective we want to underscore in this study is that using the right amount and type of data is sufficient to compete with existing solutions with no improvements to the architecture. Now that we have established a hierarchy of multispectral bands, we can further look into designing a novel architecture or training regime for water segmentation. Our process can also be extended to other features such as vegetation or built-up land and can be used in combinations for multi-class segmentation.

\subsubsection{Acknowledgment}
This work has been supported by Center of Data Science and Applied Machine Learning, Computer Science and Engineering Department of PES University, and Regional Remote Sensing Centre - south. We would like to thank Dr. Shylaja, S.S. of PES University and Dr. K. Ganesha Raj of Regional Remote Sensing Centre - south for the opportunity to carry out this work.
%
%
%
\bibliographystyle{splncs04}
\bibliography{samplepaper}
\end{document}